%% file: main.tex
\documentclass[10pt,twocolumn,letterpaper]{article}

\usepackage{wacv}
\usepackage{times}
\usepackage{epsfig}
\usepackage{graphicx}
\usepackage{amsmath}
\usepackage{amssymb}

\usepackage{algorithmic}
\usepackage{textcomp}
\usepackage{xcolor}
\usepackage{multirow}
\usepackage{color}
\usepackage{subfigure}
\usepackage{cite}

\usepackage{caption}
\usepackage{adjustbox}
\usepackage{wrapfig}

\def\wacvPaperID{0044} 

\wacvfinalcopy 

\ifwacvfinal
\def\assignedStartPage{9876} 
\fi

\ifwacvfinal
\usepackage[breaklinks=true,bookmarks=false]{hyperref}
\else
\usepackage[pagebackref=true,breaklinks=true,colorlinks,bookmarks=false]{hyperref}
\fi

\ifwacvfinal
\else
\fi

\begin{document}

\title{mToFNet: Object Anti-Spoofing with Mobile Time-of-Flight Data}
\author{Yonghyun Jeong$^{1\dagger}$, Doyeon Kim$^{1\dagger}$, Jaehyeon Lee$^1$, Minki Hong$^1$, Solbi Hwang$^1$, Jongwon Choi$^2$\thanks{Corresponding author.}\\
$^1$Samsung SDS, Seoul, Korea\\
{\tt\small  yhyun.jeong, dy31.kim, jhreplay.lee, mkidea.hong, solbi2.hwang@samsung.com}
\and
$^2$Dept. of Advanced Imaging, Chung-Ang University, Seoul, Korea\\
{\tt\small choijw@cau.ac.kr}
}

\twocolumn[{
\renewcommand\twocolumn[1][]{#1}%
\maketitle
\begin{center}
   \centering
    \vspace{-7mm}
    \includegraphics[width=\linewidth]{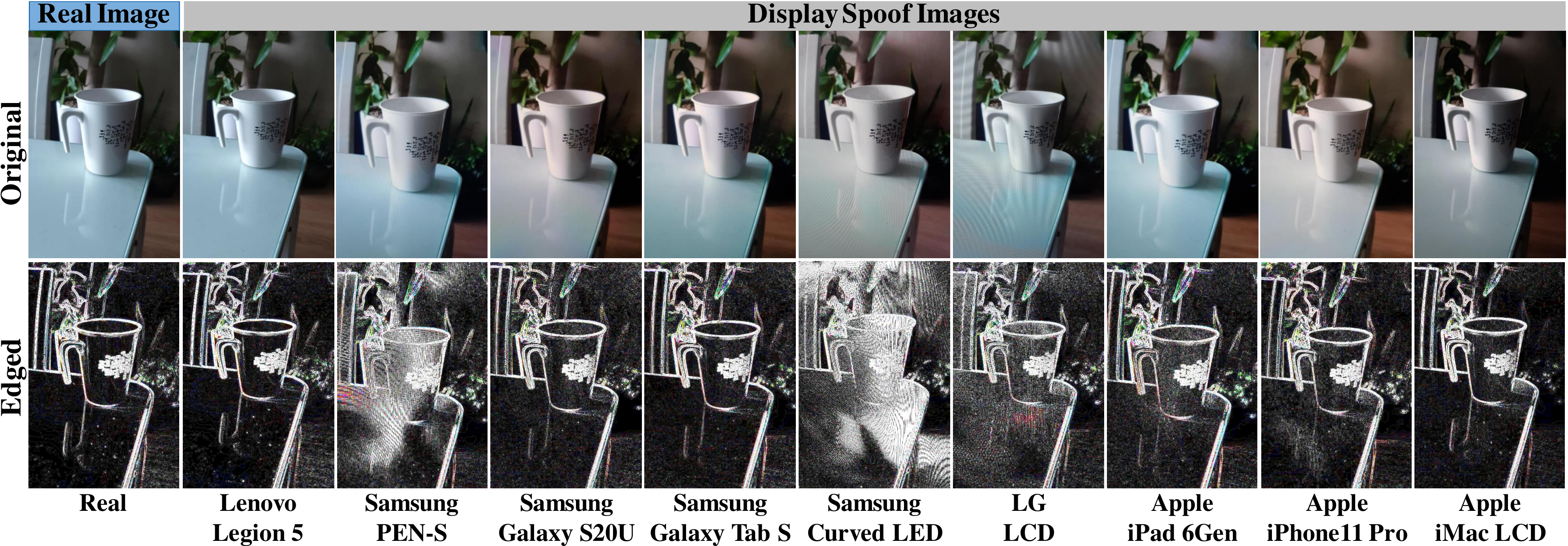}
    \captionof{figure}{\textbf{Comparison between real and display images showing diverse moiré patterns in edged images.}
    The artifacts varying by the displays are known as moiré patterns, which can lead to overfitting of the model to the training data. The unique patterns per display makes it challenging to develop a generalized method to distinguish such spoof images. 
    Since each display has a unique moire pattern, it is challenging to develop a generalized model to distinguish such spoof images. }\vspace{-2mm}
    
    \label{fig:teaser}
\end{center}
}]

\thispagestyle{empty}

\input{0.abstract}
\input{1.introduction}
\input{2.related_work}
\input{3-1.method} 
\input{3-2.dataset}

\input{4.experiments}

\input{5.conclusion}

{\small
\bibliographystyle{ieee_fullname}
\bibliography{egbib}
}


\end{document}

%% file: 0.abstract.tex
\begin{abstract}
\vspace{-3mm}
\let\thefootnote\relax\footnotetext{$\dagger~Equal~contribution.$}
\let\thefootnote\relax\footnotetext{$*~Corresponding~author.$}
In online markets, sellers can maliciously recapture others' images on display screens to utilize as spoof images, which can be challenging to distinguish in human eyes. 
To prevent such harm, we propose an anti-spoofing method using the paired rgb images and depth maps provided by the mobile camera with a Time-of-Fight sensor.
When images are recaptured on display screens, various patterns differing by the screens as known as the moiré patterns can be also captured in spoof images. These patterns lead the anti-spoofing model to be overfitted and unable to detect spoof images recaptured on unseen media.
To avoid the issue, we build a novel representation model composed of two embedding models, which can be trained without considering the recaptured images.
Also, we newly introduce \textit{mToF} dataset, the largest and most diverse object anti-spoofing dataset, and the first to utilize ToF data. Experimental results confirm that our model achieves robust generalization even across unseen domains.
\end{abstract}

%% file: 1.introduction.tex
\vspace{-5mm}
\section{Introduction}
As the volume of online transactions increases, the size of online person-to-person (P2P) transactions is also on the rise (i.e. Craigslist). 
In unfortunate cases, sellers can maliciously use spoof images for scams, and buyers are forced to bear the risk of scams to proceed with transactions.
To prevent such cases, many online services provide mobile applications specifically developed for secure verification with real-time capturing and direct transferring of users' images. 
However, such verification methods are still imperfect because the abusers can avoid such safeguards by recapturing others' images displayed on a screen. Thus, distinguishing such spoof images has become one of the most important challenges to foster reliable online transactions. 

Unfortunately, however, most previous anti-spoofing studies have focused on the human face~\cite{lbp0,lbp_top,lbp2,context_based,context_based2,dl_faceantispoof,facede_spoofing,atoum,celebA_Spoof}.
While the common characteristics of the human face can be utilized in the face anti-spoofing detector, the object anti-spoofing detector cannot utilize the architectural properties of the target objects due to their variety in merchandise.
As a result, the object anti-spoofing detector needs to focus on the difference between the real and display images, which originates from the capturing constraints.

As shown in the edged images located at the second row of Fig.~\ref{fig:teaser}, some display screens show distinct artifacts, which are known as moiré patterns.
The moiré patterns can be utilized to train the anti-spoofing model; however, it cannot detect the display images from the unseen spoof media due to the uniqueness of the moiré patterns.
When overfitting occurs, the model's performance can dramatically decline when tested with unseen data during the training phase.
Furthermore, the moiré patterns of some display screens appear in a subtle and almost indistinguishable manner in human eyes.
Thus, for the generality across the various spoof media, we need to develop a more advanced anti-spoofing detector to avoid the overfitting issue to the moiré patterns.

To solve the issues, we propose a novel framework to utilize both the image and the depth map.
The proposed framework contains dual embedding models to learn the multi-sensor representation of the real images, which are trained without the display images.
Since no display image is considered during training, the representation model can ignore the effect of moiré patterns entirely. Thus, our proposed model can achieve improved robustness across the various types of spoof media.
To train and evaluate our model, we collect \textit{mToF} dataset, which provides the largest amount of real and display images captured with the various objects on diverse spoof media, each paired with the ToF map.
The ToF map is the depth map obtained with the ToF sensor, which estimates the depth based on the duration of the light emitted from the sensor to reach the object and return.
Our mToF dataset is the largest and the most diverse object anti-spoofing dataset and the first to utilize ToF maps in this field of study. 
By using the mToF dataset, numerous experimental results confirm that our anti-spoofing method can outperform other models and achieve state-of-the-art performance on generalized detection in various combinations of objects and spoof media.  
\footnote{https://github.com/SamsungSDS-Team9/mToFNet}
\vspace{-0.5mm}
\begin{itemize}
\item In the field of object anti-spoofing, our study is the first to employ the images with depth information gathered by the mobile ToF sensor.
\vspace{-0.5mm}
\item Using the RGB images with the depth maps, we propose a generalized anti-spoofing method to distinguish even the unseen display images during the training phase.
\vspace{-0.5mm}
\item We introduce a new dataset of $12,529$ pairs of RGB images and the corresponding ToF maps, all of which are labeled as either `real' or `display.'
\vspace{-0.5mm}
\item Numerous experiments validate the effectiveness of our method in object anti-spoofing.
\end{itemize}

%% file: 2.related_work.tex
\vspace{-0.5mm}\section{Related Work}

\subsection{Face Anti-Spoofing  } 
Most previous studies in anti-spoofing methods have been focused on the face category only, usually for the biometric recognition system to allow access to genuine users and prevent identity theft.
For texture-based anti-spoofing, early studies focus on the hand-crafted feature descriptors, such as the local binary patterns~\cite{lbp0, lbp_top, lbp2}, and histogram of the oriented gradients~\cite{context_based, context_based2}. The convolutional Neural Networks (CNN) are also employed for face anti-spoofing, such as \cite{atoum}, which utilizes CNN and score fusion methods. Also, \cite{dl_faceantispoof} uses a combination of CNN, Principle Component Analysis (PCA), and Support Vector Machine (SVM) as a face anti-spoofing method.
Then, \cite{facede_spoofing} inversely decomposes a spoof face into a live face and a spoof noise for classification. Recently, \cite{celebA_Spoof} introduces a vast amount of face anti-spoofing dataset with rich annotations.

\subsection{Object Anti-Spoofing  }
To expand the scope, we explore the literature on the object anti-spoofing for a deeper analysis. 
Recently, \cite{goas} takes the issue of recaptured images on spoof media using the CNN-based framework consisted of GOGen, GODisc, and GOLab. \cite{goas} also provides GOSet, a new dataset consists of 2,849 videos captured with 7 camera sensors, 7 spoof media, and 24 objects for generic object anti-spoofing (GOAS). 
\cite{dofnet} tackles the issue by analyzing the difference in depth of field of two images, each with a different focal length.
\cite{dofnet} provides a unique paired dataset using various objects and three spoof media, and each pair consists of two images with the same viewpoint but different focal lengths.
Our work improves upon the prior literature by utilizing various object categories and spoof media with mobile ToF data, and providing the most rich contents and the largest amount of dataset. 

\subsection{Studies on RGB-D Images}
Generally referred to the colored images containing depth information, RGB-D is employed in various research areas.
\cite{2d-driven} suggests placing 3D bounding boxes to detect objects using 2D information in RGB-D images, in order to decrease the run-time and 3D search space. 
For RGB-D salient object detection, \cite{chenli} proposes a fusion method for the RGB and depth information using CNN.
Also, RGB-D is employed for object classification, as \cite{socher} introduces a model that combined convolutional and recursive neural networks.
Also, \cite{fehr, Beksi} suggest using RGB-D for object classification using dictionary learning and covariance descriptors. 
RGB-D is used in depth estimation to improve performance, as many studies based on deep networks suggest~\cite{eigen2015, liu2015, roy2016, xie2016}. 
Recently, with a spread of Time-of-Flight (ToF) sensors in mobile cameras, \cite{qiu} suggests a joint alignment and refinement using deep learning for the ToF RGB-D module.
We find that it is useful to detect the display images using the depth maps, which have not been used for anti-spoofing yet, since the conventional spoof media result in limited variations in depth maps.  

%% file: 3-1.method.tex
\begin{figure}[t]
\centering
        \subfigure[Difference in the real and display in image domain.]
        {\includegraphics[width=0.48\linewidth]{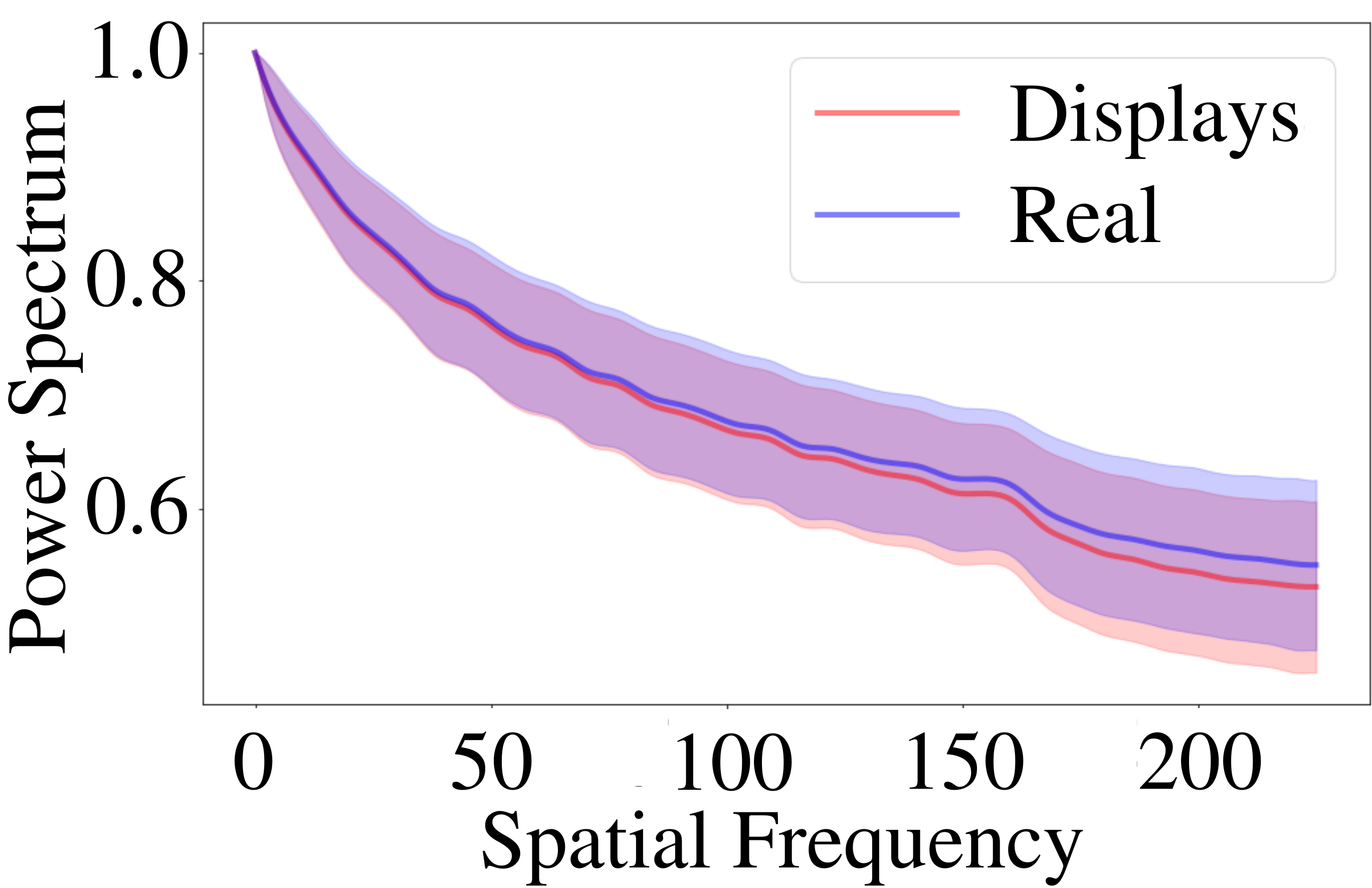}}
        \subfigure[Difference in the real and display in ToF domain.]
        {\includegraphics[width=0.48\linewidth]{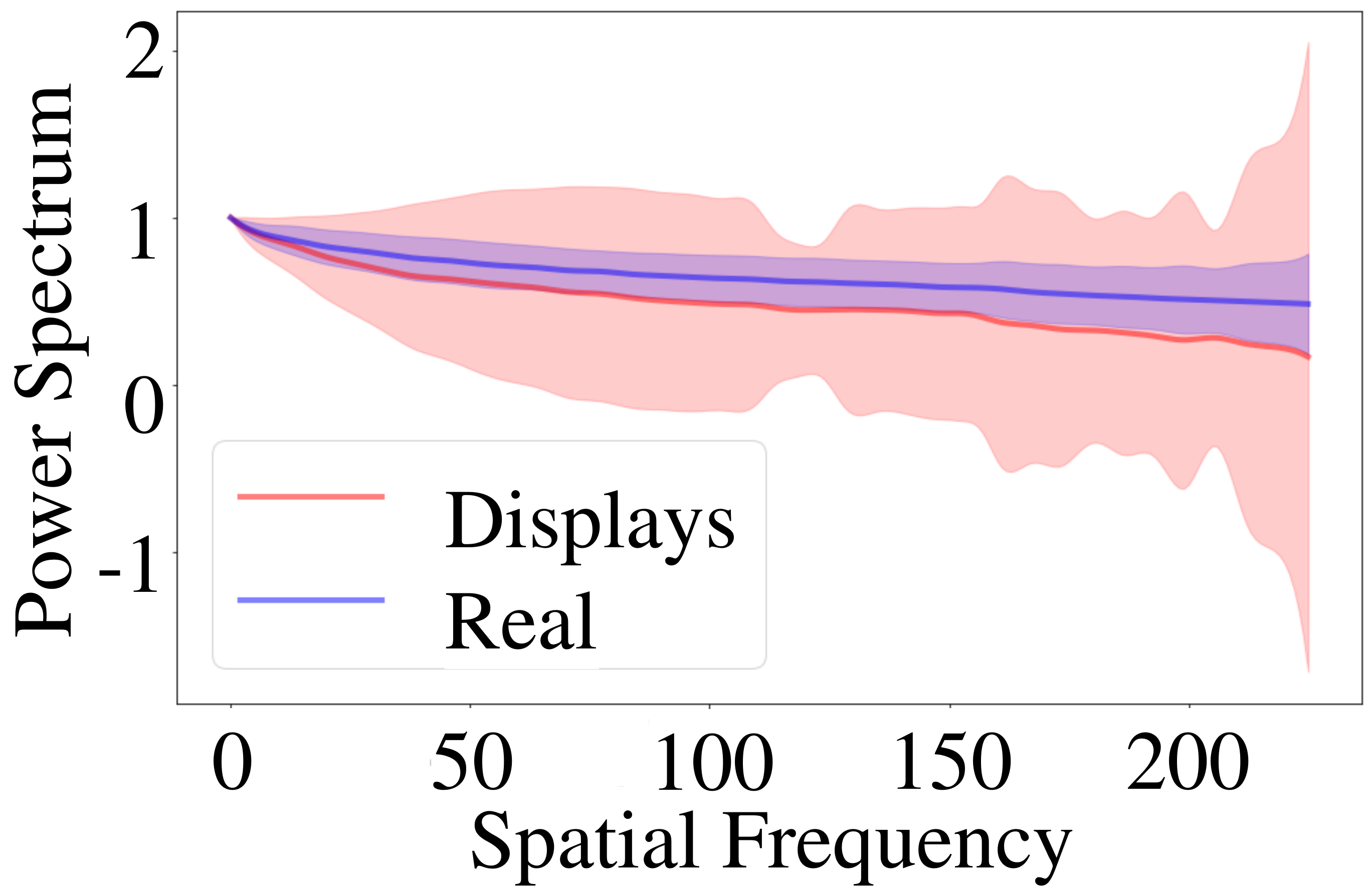}}
\caption{\textbf{Comparison in power spectra of real and display images.} 
The statistical difference between the real and display becomes magnified when ToF maps are utilized. 
}\label{fig:1d_spec}\vspace{-3mm}
\end{figure}

\vspace{-0.5mm}
\section{ToF-based Object Anti-spoofing}
\vspace{-0.5mm}
In this section, we observe the difference in ToF maps between the real pairs and display pairs, and propose a robust anti-spoofing method utilizing the ToF maps.
The real pairs of images and ToF maps are obtained by capturing the actual objects, and the display pairs are acquired by recapturing the images displayed on the screens.
First, we conduct a comparative analysis on the image level and the ToF level in Section~\ref{sec:freq_anal}, then we introduce our overall framework and its training method in detail in Section~\ref{sec:framework}.

\vspace{-0.5mm}
\subsection{ToF Frequency Analysis}\label{sec:freq_anal}
\vspace{-0.5mm}
As shown in Figure~\ref{fig:teaser}, the moiré patterns appear differently per display screen, and some of them are very subtle and difficult to distinguish in human eyes.
Compared to the moiré patterns of the display images, the ToF maps show distinct differences between the real and display, which even work across various spoof media.
To compare the characteristics of the images and ToF maps, we conduct a frequency-level analysis.
First, we transform the 2-Dimensional (2D) images and ToF maps into magnitude spectrum by applying Discrete Fourier Transform (DFT).
DFT is a mathematical approach to disintegrate a discrete signal into the frequency-level components ranging from zero up to the maximum frequency that can be represented when the spatial resolution is given~\cite{fft}. 

To reduce the dimension of the 2D spectrum from the images and ToF maps, the frequency-level 2D spectrum is transformed into 1D power spectrum by applying Azimuthal averaging~\cite{unmasking}, which is a computational approach to obtain a robust 1D representation from the FFT power spectrum.
Utilizing the method, we can scale down the number of features but maintain the relevant information. 
By using the training set of our mToF dataset, Figure~\ref{fig:1d_spec} shows the comparison between the 1D power spectrum of the images and the ToF maps. 
The detailed explanations on mToF dataset are given in Section~\ref{sec:dataset}.
As illustrated, the ToF maps contain more `artifacts' or `patterns' from the display screens and thus show a greater difference in distributions.
Based on the characteristics, we design the overall framework as described in the following section.

\begin{figure*}[t]
\centering
{\includegraphics[width=0.95\linewidth]{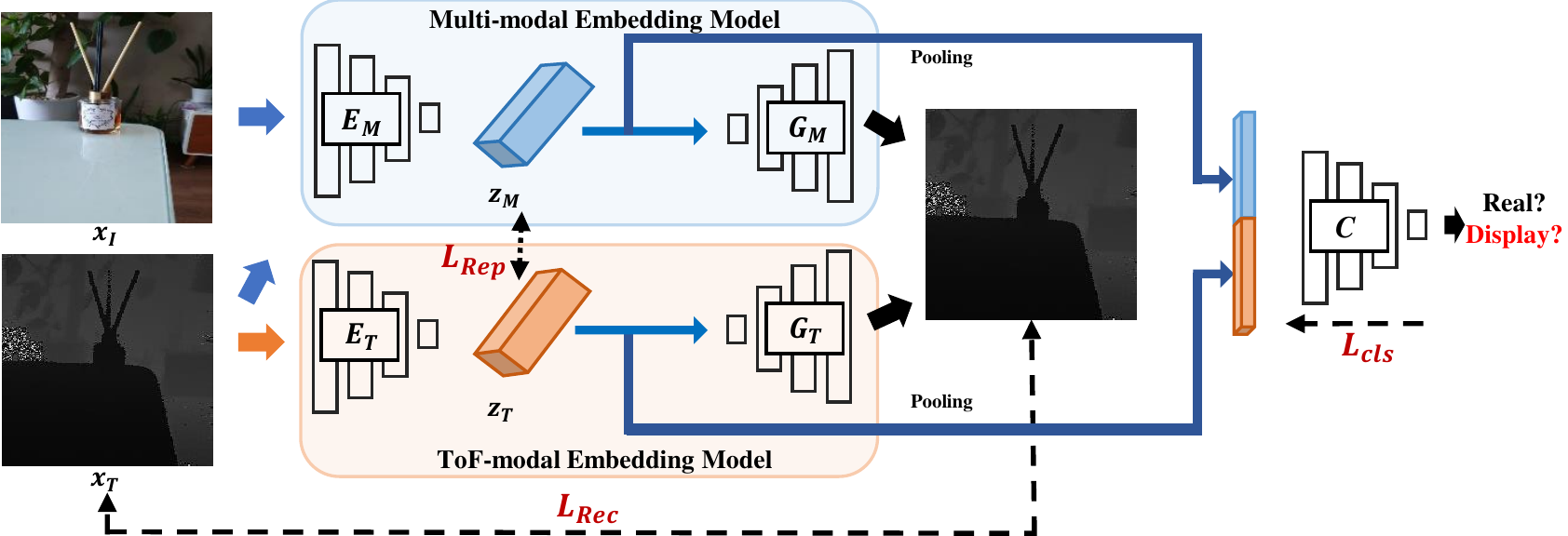}}
\caption{\textbf{Overall framework. } Using the real images and their paired ToF maps as the inputs, the first embedding model is trained to reconstruct the ToF maps. Then, using the sampled ToF maps, the second embedding model is trained to reconstruct the identical ToF maps. The representation features from the two embedding models are mapped into the identical feature space, from which the classifier makes the prediction. }
\label{fig:framework}\vspace{-5mm}
\end{figure*}

\vspace{-0.5mm}
\subsection{Overall Framework}\label{sec:framework}
\vspace{-0.5mm}
Using the two types of modalities including the images and the ToF maps, we design a framework to distinguish between the real and display pair without using the moiré patterns.
To overcome the overfitting issue by the moiré patterns of the training data, we need to entirely ignore the rgb images of the display pairs. 
Thus, we utilize the ToF maps for the display pairs, while both the images and the ToF maps are used for the real pairs. 
Since the conventional classification network cannot be trained by the inconsistent type of the input data, we build a ToF representation network that can be trained even without the display images. 

ToF representation network contains two separate embedding models trained to represent the data distributions of only the real pairs and both pairs, respectively. 
The embedding model with the real pairs is named as \textit{multi-modal embedding model}, which receives both the images and the ToF maps and reconstructs the ToF maps.
The other embedding model with both pairs is named as \textit{ToF-modal embedding model}, and it only receives the ToF maps to recover the identity ToF maps.
We let the two representation features of the embedding models be similar to each other upon the real pairs, which results in the abnormal distribution of the representation features for the display pairs. Then, the two representation features of the embedding models are concatenated to be inserted into the spoof classifier, which detects the display images by recognizing the dissimilarity of the two representation features.
The overall framework is illustrated in Figure~\ref{fig:framework}.

\subsubsection{ToF Representation Network}
We define the input image and the ToF map as $x_I\in\mathbb{R}^{w\times h\times 3}$ and $x_T\in\mathbb{R}^{w\times h}$, respectively, where $w$ is the width and $h$ means the height of data. We assume that the two types of data are resized to have the same size.
The two embedding models respectively contain an encoder and the following generator.
The encoders of the embedding models compress the input data into the representation feature, which is used to reconstruct the input ToF map by the following generator.
We define the encoder and generator of the multi-modal embedding model as $E_M$ and $G_M$, respectively, and the encoder and generator of the ToF-modal embedding model are denoted as $E_T$ and $G_T$, respectively.

$E_M$ is inserted by the 4-channel feature $c(x_I, x_T)$ concatenating $x_I$ and $x_T$ of the real pairs.
Then, the generated latent code is defined by $z_M$ (\textit{i.e.} $E_M(c(x_I, x_T))$). 
From the generated $z_M$, the following $G_M$ reconstructs the input ToF map.
Instead of the concatenated features, only the ToF map is reconstructed by $G_M$, which leads to stable training of the embedding model by reducing the information to be reconstructed.
Unlike the multi-modal embedding model ($G_M(E_M(\bullet))$), the ToF-modal embedding model ignores the images as its input.
Thus, $E_T$ gets the 1-channel feature of $x_T$, and its generated latent code can be obtained as $z_T\equiv E_T(x_T)$.
Finally, the generator $G_T$ of the ToF-modal embedding model ($G_T(E_T(\bullet))$) reconstructs the input ToF map by estimating $G_T(z_T)$.
The encoders contain three convolution layers of stride 2 and kernel size 3, which are respectively followed by a ReLU activation function~\cite{relu} and a batch normalization.
In the generators, three transposed convolution layers are serially connected with stride 2 and kernel size 2.
To allow the representation feature to contain the data distribution, we do not employ the U-Net architecture~\cite{pix2pix_gan} nor the skip connection~\cite{resnet} for our embedding models.

The training loss for the ToF representation network consists of two reconstruction losses respectively for each embedding model and one representation loss.
Using the reconstruction losses, we train the two embedding models to effectively represent the ToF information.
The multi-modal embedding model is trained to always reconstruct the input ToF maps only from the real pairs.
Thus, for the training phase of the multi-modal embedding model, the input images are sampled as $(x_I,x_T)\sim X_{real}$, where $X_{real}$ is the set of real pairs.
The reconstruction loss for the multi-modal embedding model can be defined as follows: 
\begin{equation}\label{eq:rec_loss}
        \mathcal{L}_{rec}^{M}=\mathbb{E}_{(x_I, x_T)\sim X_{real}}[||x_T-D_M(E_M(c(x_I,x_T)))||_{2}].
\end{equation}
In the case of the ToF-modal embedding model, its encoder receives only the ToF map, while its generator reconstructs the ToF map similarly to that of the multi-modal embedding model. 
Thus, the reconstruction loss for the ToF-modal embedding model can be defined as follows: 
\begin{equation}\label{eq:rec_loss}
        \mathcal{L}_{rec}^{T}=\mathbb{E}_{(x_I, x_T)\sim X}[||x_T-G_{T}(E_{T}(x_T))||_{2}],
\end{equation}
where $X$ presents the set of all pairs.

In addition, to detect the display pairs by using the representation features, we design a representation loss to reduce the distance of the representation features from the two embedding models when the real pairs are given.
By applying the loss, only the representation features of real pairs are similar to each other, so the display pairs can be easily detected based on their discrepancies.
Then, the representation loss can be derived as follows:
\begin{equation}\label{eq:rec_loss}
        \mathcal{L}_{rep}=\mathbb{E}_{(x_I, x_T)\sim X_{real}}[||E_{M}(c(x_I, x_T)) - E_{T}(x_T)||_{1}].
\end{equation}

\subsubsection{Spoof Classifier}
The spoof classifier compares the representation features from the two embedding models, based on the similarity between the latent codes to predict the real and display pairs.
At first, $z_M$ and $z_I$ are pooled by an average pooling and vectorized, which are defined by $\hat{z}_M$ and $\hat{z}_I$, respectively.
Then, the spoof classifier is fed by the concatenated feature of $[\hat{z}_M, \hat{z}_I]$ using $\hat{z}_M$ and $\hat{z}_I$.
The spoof classifier consists of multiple fully-connected layers to classify the concatenated features as one of the real and display pairs.
To operate the spoof classifier, only the encoders of the two embedding models are necessary, so we remove their generators to improve the computational efficiency in the test phase.
The spoof classifier is trained for the binary classification of `real' or `display', so we utilize the conventional cross-entropy loss with softmax function~\cite{cox1958regression}.

%% file: 3-2.dataset.tex
\section{mToF Dataset}\label{sec:dataset}

\input{tables/dataset_comparison}

\input{tables/dataset_displays3}

Recently, several mobile manufacturers have begun to equip ToF sensors on their mobile devices such as Samsung Galaxy S20+, Apple iPhone 12 Pro, and LG G8 ThinkQ. Due to its easy accessibility and simple operation, the ToF sensor is a great tool for the measurement of depth information. 
Our \textit{mToF} dataset is collected to overcome the limitations in size and variety of the previous object anti-spoofing datasets and to provide additional ToF data for the first time in this area of research.
With 12,529 images in 27 categories captured on 16 different spoof media, mToF dataset is the largest in size with the most variety, compared to other recent anti-spoofing datasets as shown in Table~\ref{tab:dataset_comparison}.
Using the ToF map, we can effectively distinguish whether an image is taken of a real object, or recaptured on a display medium. 
Since most anti-spoofing studies are conducted in the face category~\cite{lbp0, lbp_top, lbp2, context_based, context_based2, atoum, dl_faceantispoof, facede_spoofing, celebA_Spoof}, the non-facial datasets are scarce~\cite{goas, dofnet}.
Thus, we expect mToF dataset to bring a meaningful contribution in the field of object anti-spoofing. 

\vspace{-0.5mm} \subsection{Data Composition }
\vspace{-0.5mm}
Our newly introduced mToF dataset can be divided into two major segments: the images taken of the real objects are defined as \textit{real images}, and the recaptured images of the real images on the spoof media are defined as \textit{display images}. Each real and display image is paired with its corresponding ToF data, which is a unique feature compared to other datasets. 
For data diversity, our mToF dataset is composed of 27 object categories, including book, bottle, bowl, bug spray, candle, cellphone holder, condiment, cosmetic, cup, diffuser, dish, food container, glasses case, household goods, humidifier, mouth wash, music album, ointment, pan, perfume, pot, snack, toy, vitamin, wallet, wet-wipe, and window cleaner.

Also, we use 16 different spoof media, including various monitors, laptops, mobile phones, tablet PCs, and projectors from diverse manufacturers, as listed in Table~\ref{tab:dataset_display3}.
For monitors, four types of display screens are used, including a Samsung Wide Quad High Definition (WQHD) Light-Emitting Diode (LED) monitor (S27H850QF, 27-inch, 2019), a Samsung WQHD curved quantum-dot LED monitor (C34J791WT, 34-inch, 2018), a retina 5K liquid-crystal display (LCD) of Apple iMac (A1419, 27-inch, 2017), and an LG anti-glare LCD monitor (32-inch, 32GK850G, 2017). 
Also, four types of laptops are used, including a Samsung PEN-S (NT930, 13.3-inch, 2019), a Lenovo Legion 5 (Y540, 15.6-inch, 2020), an Apple Macbook Pro (A1398, 15.4-inch, 2015), and an LG Gram (15Z990, 15.6-inch, 2019). 
For mobile phones, four types of devices are used, including a Samsung Galaxy S20 Ultra (SM-G988, 6.9-inch, 2020), a Samsung Galaxy Note 8 (SM-N950N, 6.3-inch, 2017), an Apple iPhone 12 Pro (A2407, 6.1-inch, 2020), an Apple iPhone 11 Pro (A2215, 5.8-inch, 2019). 
Lastly, two types of projectors are used, including a projector screen for NEC LCD projector (NP-M311XG, 2012) and an LG Cinebeam Digital Light Processing (DLP) projector(HF60LA, 2019).

\vspace{-0.5mm}\subsection{Data Collection  }
\vspace{-0.5mm}
All images are captured with a mobile application specifically developed to obtain a pair of RGB images and its corresponding ToF maps, using the mobile ToF sensor on a Samsung Galaxy Note 10.
For data collection, we first take the real images by focusing on the subjects located in the middle and then recapture the display images by setting the mobile phone on a tripod in front of the screens displaying the real pictures. 
All display images are taken in the dark with the lights off to minimize unnecessary factors, such as light reflections and noises.
Also, to imitate real-world settings, we include various backgrounds behind the subjects to provide a variety of shooting environments. 
The backgrounds include bookshelves, curtains, home appliances, kitchen cabinets, paintings, and more. 
For training, the images are randomly mixed to prevent data bias.

\vspace{-0.5mm}
\subsection{On-device Refinement of ToF Maps}
\vspace{-0.5mm}
The raw ToF maps captured with the ToF sensor require a refining process due to the numerous artifacts and noises included, which are affected by various factors, such as multi-path interference, motion artifacts, shot noises, and different camera response functions \cite{guo, qiu}.
After capturing the paired images with the ToF maps, we refine the data within the mobile application according to the API guide provided by Android\footnote{https://developer.android.com/reference/android/graphics/ImageFormat}.
For refinement, we acquire the ToF map with an even width and height, in which each pixel is $16$-bit valued indicating the range of ToF measurements. 
When the ToF map is captured, the correctness of the depth value is saved along with the depth value at the three most significant bits of them.
The correctness of the depth value is encoded in the three bits as follows: when a value of zero represents 100\% confidence, one represents 0\% confidence.
Our collected images have a resolution of 1280$\times$720, and their ToF maps have a resolution of 240$\times$180.

\vspace{-0.5mm}
\subsection{Pre-processing of Paired Images}
\vspace{-0.5mm}
Without pre-processing, we cannot use the RGB image and the depth map simultaneously, since the depth maps from the ToF sensor is represented by 16 bits while a color channel of the RGB images is in 8 bits.
To equalize the bit scales of the RGB images and the depth maps, we transform the bit scale of the depth map into 8 bits where the value ranges between 0 and 255.
Thus, the bit scale of the depth map becomes equivalent to one color channel of the RGB image.
Afterward, we resize all of the RGB images and the depth maps to $180\times 180$ in resolution, which is the smallest image length among the captured data.
Finally, when the samples captured with the same target object and the same type of displays are grouped by a sample set, we compose each sample set as follows: $80\%$ as training set, $10\%$ as validation set, and the last $10\%$ as test set. The sample sets of real images are also organized in the same fashion.

%% file: tables/dataset_comparison.tex
\begin{table}[]
\centering
\scriptsize
\caption{Dataset Comparison}\vspace{-2mm}
\label{tab:dataset_comparison}
\resizebox{1.00\linewidth}{!}{%
\begin{tabular}{lccrrrcc}
\hline
Dataset & Category & Domain & \multicolumn{1}{c}{Size} & \begin{tabular}[c]{@{}c@{}}\# of Spoof\\ Subject\end{tabular} & \begin{tabular}[c]{@{}c@{}}\# of Spoof\\ Medium\end{tabular} & Paired & ToF \\ \hline
Celeb-A Spoof\cite{celebA_Spoof} & Face & Image  & 625,537 & 10,177 & - & No & No \\
GOAS~\cite{goas} & Object & Video  & 2,849  & 24 & 7 & No & No \\
DoFNet~\cite{dofnet} & Object & Image  & 2,757  & 6 & 3 & Yes & No \\
Ours & Object & Image  & 12,529 & 27 & 16 & Yes & Yes \\ \hline
\end{tabular}
}

\end{table}

%% file: tables/dataset_displays3.tex
\begin{table}[]
\centering
\scriptsize
\caption{Types of Spoof Mediums for mToF Dataset}\vspace{-2mm}
\label{tab:dataset_display3}
\resizebox{1.00\linewidth}{!}{%
    \begin{tabular}{ccccc} 
    \hline
    \textbf{Monitor} & \textbf{Laptop} & \textbf{Mobile Phone} & \textbf{Tablet PC} & \textbf{Projector}                                             \\ 
    \hline
    \begin{tabular}[c]{@{}c@{}}Samsung LED\\ (S27H850QF)\end{tabular} & \begin{tabular}[c]{@{}c@{}}Samsung PEN-S\\ (NT930)\end{tabular}     & \begin{tabular}[c]{@{}c@{}}Samsung S20 Ultra\\ (SM-G988)\end{tabular} & \begin{tabular}[c]{@{}c@{}}Samsung Tab S4\\(SM-T830)\end{tabular} & \begin{tabular}[c]{@{}c@{}}NEC LCD\\ (M311XG)\end{tabular}         \\ 
    \hline
    \begin{tabular}[c]{@{}c@{}}Samsung Curved LED\\ (C34J791WT)\end{tabular}    & \begin{tabular}[c]{@{}c@{}}Lenovo Legion 5\\ (Y540)\end{tabular}    & \begin{tabular}[c]{@{}c@{}}Samsung Note 8\\ (SM-N950N)\end{tabular}   & \begin{tabular}[c]{@{}c@{}}Apple iPad 6Gen\\(A1893)\end{tabular}  & \begin{tabular}[c]{@{}c@{}}LG CineBeam DLP\\(HF60LA)\end{tabular}  \\ 
    \hline
    \begin{tabular}[c]{@{}c@{}}Apple iMac LCD\\ (A1419)\end{tabular}     & \begin{tabular}[c]{@{}c@{}}Apple Macbook Pro\\ (A1398)\end{tabular} & \begin{tabular}[c]{@{}c@{}}Apple iPhone12 Pro\\ (A2407)\end{tabular}  & -                                            & -          \\ 
    \hline
    \begin{tabular}[c]{@{}c@{}}LG LCD\\ (32GK850G)\end{tabular}          & \begin{tabular}[c]{@{}c@{}}LG Gram\\ (15Z990)\end{tabular}          & \begin{tabular}[c]{@{}c@{}}Apple iPhone11 Pro\\(A2215)\end{tabular}   & -                                            & -          \\
    \hline
    \end{tabular}
}

\end{table}

%% file: 4.experiments.tex
\input{tables/patial_results}

\input{tables/ablation}

\input{tables/nums_results}

\vspace{-0.5mm}
\section{Experimental Results}
\vspace{-0.5mm}
In this section, we conduct experiments to evaluate our model in various scenarios.
The implementation details of our method are as follows. To align the resolutions of the image and ToF domains, we resize the image resolution to $180\times 240$ to match the resolution of ToF data. For data augmentation, we randomly crop the image size to $160\times 160$. 
The networks are trained based on Adam optimizer~\cite{adam} using the learning rate of 0.0001 and the batch size of 32 with 20 epochs. 
All experiments are conducted using a single NVIDIA Titan RTX GPU.
To effectively evaluate the performance of our method, we employ the measurements commonly used in anti-spoofing and image forgery detection:
accuracy (Acc.), AUROC, and average precision (A.P.)~\cite{cozzolino, zhang, adobe}. 

\vspace{-0.5mm}
\subsection{Comparison models}\label{sec:comparison_model}
\vspace{-0.5mm}
In this section, we provide explanations for the compared models including the PCA-based, frequency-based, and CNN-based naive models.

\vspace{-0.5mm}
\subsubsection{PCA-based Model} 
\vspace{-0.5mm}
To find a better representation method for detecting display images than the variance of ToF maps, we employ the Principal Component Analysis~(PCA)~\cite{pca}, which is one of the most popular methods to find the representative features from raw data.
Using PCA, we can obtain the projection axis maximizing the discrimination of the input samples, which is called \textit{the principal component}.
Also through PCA, we can get the multiple principal components that maximize the discrimination, however the principal components are constrained to be orthogonal to each other.
After reducing the feature size of the ToF maps by two, we integrate a linear Support Vector Machine~(SVM) to detect the display pairs by using them.

\vspace{-0.5mm}
\subsubsection{Frequency-based Model}\vspace{-0.5mm}
To show the effectiveness of the frequency-based detector with the ToF maps, we also employ a frequency-based detector~\cite{watch_cvpr20} as one of the compared algorithms.
In the detector, 2-D image or ToF map is transformed into the frequency domain by Fast Fourier Transform~\cite{fft}, first. Then, by utilizing the operation of azimuthal average~\cite{unmasking}, the 2-D frequency domain is compressed into a 1-D power spectrum that compresses the frequency information of the ToF maps.
The classification model is based on SVM~\cite{svm} with linear kernel, which is also employed for the proposed framework.

\vspace{-0.5mm}
\subsubsection{CNN-based Naive Classifier}\vspace{-0.5mm}
For the classification model of Convolution Neural Networks (CNN), we utilize ResNet~\cite{resnet}, VGG~\cite{vgg}, resnext~\cite{resnext}, alexnet~\cite{alexnet}.
In a comparative analysis, we concatenate the ToF maps and the images to use them as the training input.
Since CNN simultaneously works as the feature extractor, we utilize the raw ToF maps for CNN instead of the feature extraction methods. 
We replace the last fully-connected layer of the models to reduce the number of classes by $2$.
Then, the network is fine-tuned by using the softmax cross-entropy loss for the binary classification.
The network is updated by Adam optimizer~\cite{adam}, and the number of epochs, the learning rate, and the batch size are set to $20$, $0.0001$, and $32$, respectively.

\vspace{-0.5mm}
\subsubsection{Face anti-spoofing model}\vspace{-0.5mm}

To compare our method to the face anti-spoofing model, we utilize the methodology proposed by Moon~\etal~\cite{moon2021face} and George and Marcel~\cite{georgecvpr2021}. 
Moon~\etal~\cite{moon2021face} detects the spoofing face by only using the rgb images, so the decline in performance is dramatic compared to our model.
George and Marcel~\cite{georgecvpr2021} utilizes the rgb and depth maps simultaneously like our method, and shows state-of-the-art performance in the face anti-spoofing problem.
Since only the trained model and the loss are shared publicly for the method, we implement the code for training and testing, which is fully validated by using the uploaded model.

\vspace{-0.5mm}
\subsection{Unseen Display Performance}\label{sec:unseen_result}\vspace{-0.5mm}
For real-world applications, the anti-spoofing task should be able to cover the unseen environment, because it is impractical to gather all necessary data to detect every spoof medium~\cite{goas, dofnet, adobe}.
Thus, for practical applications, we validate the proposed method by estimating the robustness in anti-spoofing when only a limited number of displays is considered during training. Table~\ref{tab:result2} presents the robustness to the unseen media when only one display is considered in the training phase.
`Target-display' experiments are of the models trained and tested using the same display types. Also, `Unseen-display' experiments are of the models tested with unseen display types, while `All-display' experiments are of the models tested with all display types.

We compare with not only the CNN-based methods~\cite{resnet,resnext,alexnet,vgg}, such as AlexNet, VGG, and ResNet, but also the PCA-based methods~\cite{dl_faceantispoof}. 
As employed in \cite{watch_cvpr20,unmasking}, we additionally compare with the frequency-based detection methods. 
As shown in the experimental results, our method achieves the most robust performance in distinguishing the display images, using the ToF maps.
Furthermore, the state-of-the-art method for face anti-spoofing detection~\cite{georgecvpr2021} shows a decline in robustness for the unseen media, even though the method also considers the depth map. This result validates that the object anti-spoofing detection cannot be solved just by applying the face anti-spoofing detectors. 
Since the moiré is easily distinguishable by every compared model except for AlexNet, the target-display performance is consistently superior among all models.
However, in the case of unseen-display experiments, our proposed framework shows state-of-the-art performance by far.

\vspace{-0.5mm}
\subsection{Ablation}\label{sec:ablation}\vspace{-0.5mm}
Table~\ref{tab:ablation} indicates the experimental results of the ablation study to validate the individual component of our method.
First, we conduct experiments of our model using the CNN classifier and the images only, without the ToF maps (\textit{w/o ToF}). 
In this case, the model learns the moiré patterns~\cite{moire} of the display images for classification, which results in a decline in performance. 
Such results demonstrate the importance of using the ToF maps for accurate classification of the real and display images.
Second, we add the ToF maps along with the images as the input of the CNN classifier of our model but eliminate the representation network (\textit{w/o Representation Network}). 
By considering the ToF maps, we can achieve improved performance in distinguishing the real and display images. 
Lastly, we conduct experiments of our model without the representation loss (\textit{w/o} $L_{rep}$), which makes both the encoder and generator to exist in the same representation space (\textit{w/o $L_{ref}$}). 
Although both of the embedding models are trained to reconstruct ToF maps, the results validate the representation loss is essential since each network is independent of each other.

\begin{figure*}[t]
\centering
{\includegraphics[width=0.87\linewidth]{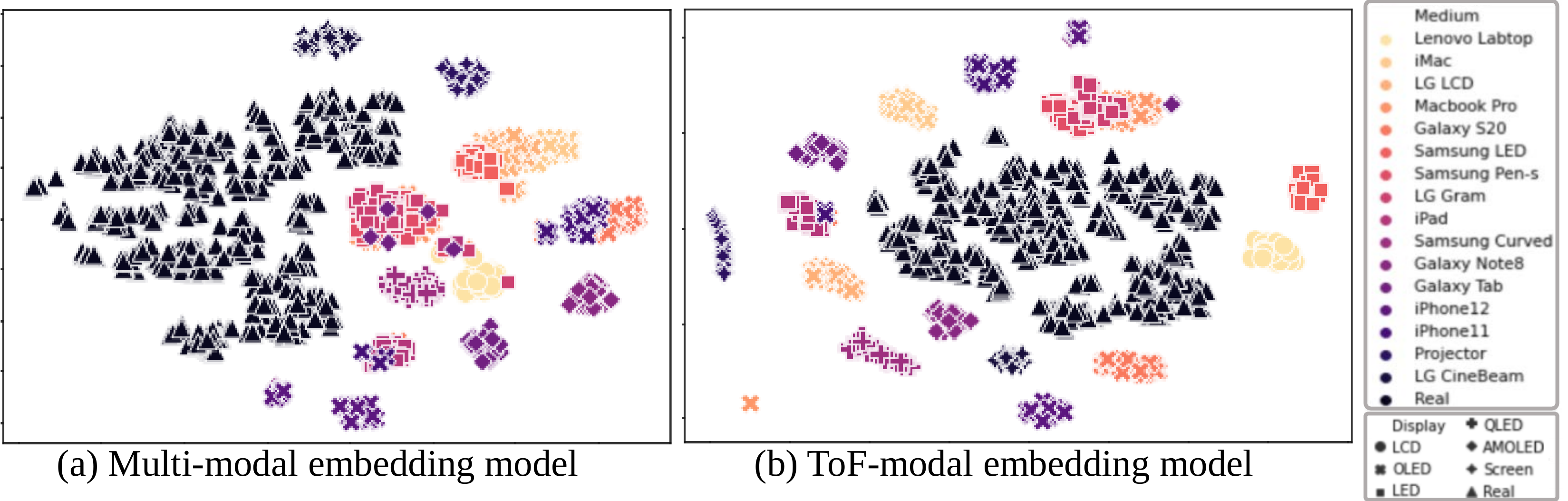}}
\caption{\textbf{t-SNE Visualization for Real and Display Pairs.} The representation features from real and display pairs are visualized by t-SNE. While the real pairs are gathered to form a stable distribution, the representation features from the display images are grouped by the spoof media and scattered out of the distribution of real pairs.}
    \label{fig:visual}\vspace{-3mm}
\end{figure*}

\begin{figure}[t]
\centering
{\includegraphics[width=1.0\linewidth]{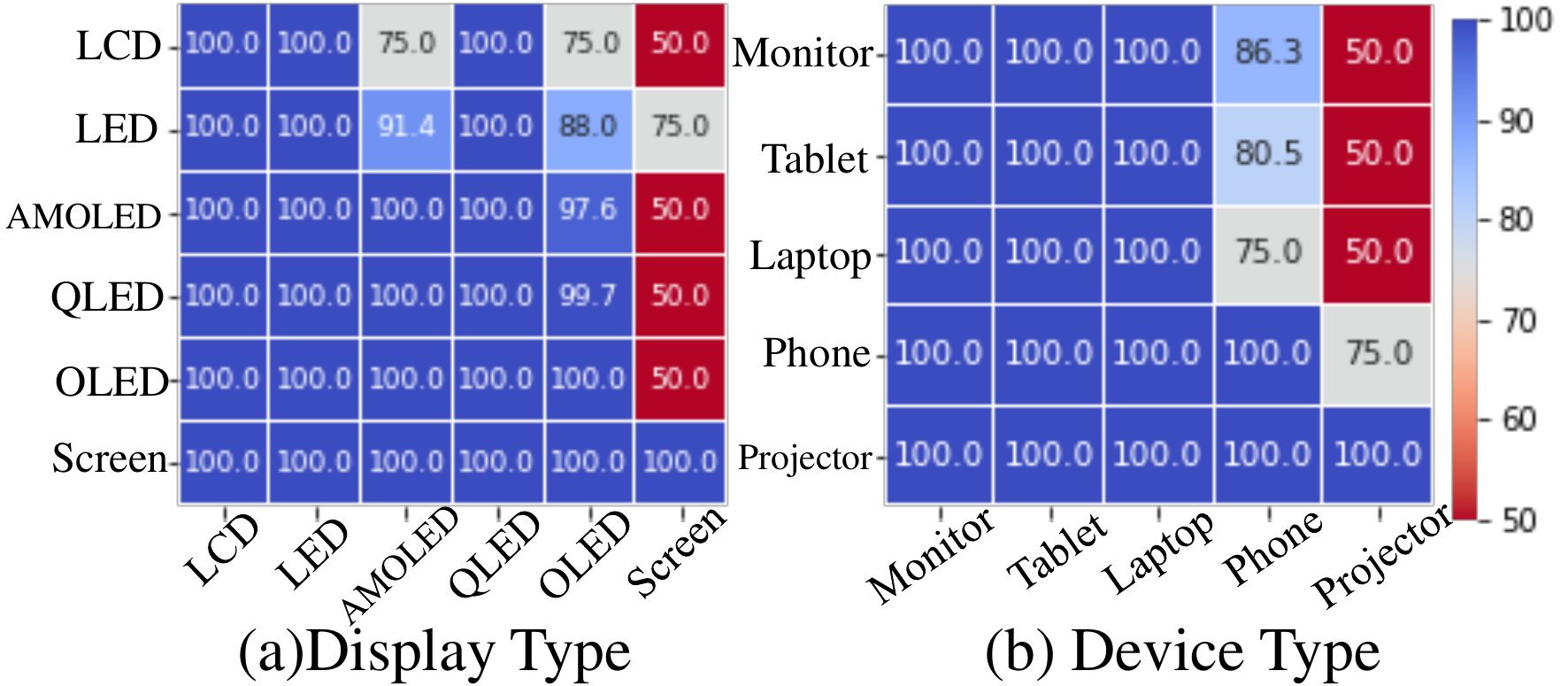}}
\caption{\textbf{Confusion Matrix. } Experiments are conducted with various combinations of training and test domains to assess our model by categorizing the spoof media by the (a) display type and (b) device type. The row and column indicate the training domain and the test domain, respectively.}
    \label{fig:confusion}\vspace{-3mm}
\end{figure}

\vspace{-0.5mm}
\subsection{Effectiveness to Train Various Moiré Patterns}\label{sec:additional_test}\vspace{-0.5mm}
In this section, we conduct additional experiments verifying that the training of numerous moiré patterns is ineffective to improve the generality of the anti-spoof detector.
For the experiments, we utilize Resnet-50~\cite{resnet}, and its hyperparameters for training are also the same as those in Section~\ref{sec:unseen_result}. 
For the compared models, to maintain the moiré patterns clearly on the display images, we randomly crop the display images to use as the input, instead of resizing as proceeded in the ToF maps. 
Also, for data augmentation, we apply random flip and random rotation of the images to expand the training data. 
By gradually increasing the number of training displays, we observe the model's performance when tested with the target-display and unseen-display. 
The experimental results are listed in Table~\ref{tab:moire}, which indicates the improved performance as the number of training displays increases. 
However, our method using the ToF maps outperforms this method using the moiré patterns. Thus, the training of numerous types of moiré is less effective than our proposed method entirely ignoring the display images.

\vspace{-0.5mm}
\subsection{Visualization of Representation Features}\label{sec:visualization}\vspace{-1.5mm}
Using t-Stochastic Neighbor Embedding (t-SNE)~\cite{tsne}, we observe whether our method based on the representation training actually makes the real images and the real ToF maps exist in the same space, so that it can effectively distinguish between the real and display images. 
Figure~\ref{fig:visual} illustrates the clear division between the feature vectors of the real and display images.
The features are well categorized according to the device type, which can provide interesting relations among the various spoofing media.

\vspace{-0.5mm}
\subsection{Analysis of Display and Device Types} \vspace{-0.5mm}
To show the different characteristics of various displays, we perform additional experiments to evaluate the performance with various combinations of training and test domains.
As shown in Figure~\ref{fig:confusion}, we build a taxonomy of the various spoof media categorized into two categories: the display types and device types.
As shown in the display types of the taxonomy, our method experiences difficulties with the screen type when trained with other media.
Similarly, in the device type, the projector type is more challenging to detect using our method trained with other media.
Also, while the characteristics of the monitor, tablet, and laptop types can be trained by the phone display, it is more challenging to detect the phone display using those characteristics only. This indicates that the robustness among the media can be related asymmetrically.
From the results, we can discover that the proposed algorithm can be improved for future investigations to enhance the robustness for the projector type and to analyze the asymmetric relations among the spoof media.

%% file: tables/patial_results.tex
\begin{table*}[]
\centering
\scriptsize
\caption{\textbf{Unseen display performance.}}\vspace{-2mm}
\label{tab:result2}
\resizebox{0.7\linewidth}{!}{%
\begin{tabular}{lcccccccccc}
\hline
\multirow{2}{*}{Models}  & \multirow{2}{*}{Data Type} & \multicolumn{3}{c}{Target-display} & \multicolumn{3}{c}{Unseen-display} & \multicolumn{3}{c}{All-display}  \\ \cline{3-11} 
   & & Acc.      & A.P. & AUROC     & Acc.      & A.P.    & AUROC  & Acc.     & A.P.   & AUROC \\ \hline
AlexNet~\cite{alexnet}& RGB, ToF & 51.37 & 88.64 & 93.59 & 40.91 & 68.77 & 62.70 & 41.56 & 70.01 & 64.63  \\
VGG~\cite{vgg} & RGB, ToF & 98.63 & 100.00 & 100.00 & 87.12 & 92.97 & 88.16 & 87.84 & 93.41 & 88.90  \\
ResNext~\cite{resnext} & RGB, ToF & 100.00 & 100.00 & 100.00 & 86.26 & 86.17 & 74.64 & 87.12 & 87.03 & 76.23  \\
ResNet~\cite{resnet} & RGB, ToF & 100.00 & 100.00 & 100.00 & 86.67 & 88.95 & 80.53 & 87.50 & 89.64 & 81.75 \\
PCA~\cite{pca} & ToF        &  100.00 & 100.00 & 100.00 & 87.31 & 87.31 & 87.31 & 88.10 & 88.10 & 88.10 \\
Moon~\etal~\cite{moon2021face} & RGB  & 73.57 & 89.75 & 85.08 & 63.33 & 75.34 & 73.80 & 63.97 & 76.24 & 74.50  \\
George \& Marcel~\cite{georgecvpr2021} & RGB, ToF  & 100.00 & 100.00 & 100.00 & 86.67 & 91.13 & 86.30 & 87.50 & 91.68 & 87.16 \\
Durall~\etal~\cite{watch_cvpr20}& ToF & 100.00 & 100.00 & 100.00 & 93.33 & 93.33 & 93.33 & 93.75 & 93.75 & 93.75\\
Durall~\etal~\cite{watch_cvpr20} & RGB, ToF  & 100.00 & 100.00 & 100.00 & 93.33 & 93.33 & 93.33 & 93.75 & 93.75 & 93.75\\
Ours  & RGB, ToF  &  100.00 & 100.00 & 100.00 & 96.67 & 100.00 & 100.00 & 96.88 & 100.00 & 100.00  \\ 
\hline
\end{tabular}
}

\end{table*}

%% file: tables/ablation.tex
\begin{table}[]
\centering
\scriptsize
\caption{\textbf{Ablation Study.}}\vspace{-2mm}
\label{tab:ablation}
\resizebox{1.0\linewidth}{!}{%
\begin{tabular}{lccccccccc}
\hline
\multirow{2}{*}{}   & \multicolumn{3}{c}{Target-display} & \multicolumn{3}{c}{Unseen-display} & \multicolumn{3}{c}{All-display}  \\ \cline{2-10} 
   & Acc.      & A.P. & AUROC     & Acc.      & A.P.    & AUROC  & Acc.     & A.P.   & AUROC \\ \hline
w/o ToF                     & 50.00 & 64.54 & 56.54 & 44.75 & 40.33 & 23.06 & 45.08 & 41.84 & 25.15  \\
w/o Rep.  & 100.00 & 100.00 & 100.00 & 87.26 & 95.94 & 92.40 & 88.06 & 96.19 & 92.88  \\
w/o $L_{rep}$               & 100.00 & 100.00 & 100.00 & 86.67 & 95.80 & 95.13 & 87.50 & 96.06 & 95.43  \\ 
Ours                        & 100.00 & 100.00 & 100.00 & 96.67 & 100.00 & 100.00 & 96.88 & 100.00 & 100.00  \\ 
\hline
\vspace{-3mm}
\end{tabular}
}

\end{table}

%% file: tables/nums_results.tex
\begin{table}[]
\centering
\scriptsize
\caption{\textbf{Moire-based Train performance.}}\vspace{-2mm}
\label{tab:moire}
\resizebox{1.0\linewidth}{!}{%
\begin{tabular}{cccccccccc}
\hline
\multicolumn{1}{c}{\multirow{2}{*}{\begin{tabular}[c]{@{}c@{}}Number of \\ Displays\end{tabular}}}   & \multicolumn{3}{c}{Target-display} & \multicolumn{3}{c}{Unseen-display} & \multicolumn{3}{c}{All-display}  \\ \cline{2-10} 
   & Acc.      & A.P. & AUROC     & Acc.      & A.P.    & AUROC  & Acc.     & A.P.   & AUROC \\ \hline
1 & 50.00 & 56.64 & 50.83 & 46.03 & 35.94 & 19.18 & 46.28 & 37.23 & 21.16 \\
2 & 50.00 & 68.88 & 66.76 & 47.90 & 39.99 & 24.96 & 48.16 & 43.60 & 30.19  \\
4 & 63.36 & 76.27 & 66.87 & 58.61 & 69.73 & 60.12 & 59.80 & 71.37 & 61.81  \\
8 & 69.78 & 87.07 & 84.36 & 64.81 & 81.57 & 74.79 & 67.30 & 84.32 & 79.58  \\
15 & 77.40 & 82.52 & 84.33 & 74.70 & 80.45 & 81.17 & 77.23 & 82.39 & 84.13  \\
Ours & 100.00 & 100.00 & 100.00 & 96.67 & 100.00 & 100.00 & 96.88 & 100.00 & 100.00  \\ 
\hline
\vspace{-3mm}
\end{tabular}
}

\end{table}

%% file: 5.conclusion.tex
\vspace{-0.5mm}\section{Conclusion}\vspace{-0.5mm}
With the expansion of online commercial transactions, it becomes increasingly important to prevent image spoofing in various categories. 
Our newly proposed method achieves the most robust performance in distinguishing the real and display images using the ToF maps, even when tested with unseen displays during the training phase.     
Numerous experiments confirm our model's robustness compared to others, and the individual components of our framework are evaluated through vigorous ablation study. 
Also, our \textit{mToF} dataset is the largest and the most diverse dataset for object anti-spoofing and is composed of the real and display images paired with the ToF maps.  
We expect our mToF dataset to be utilized in various tasks, such as 3D reconstruction and object detection. 
Furthermore, we believe our work can make ethical impacts in society by recovering the digital mistrust in online markets. 
To enhance the applicability, we plan to extend the proposed framework to work with the camera sensor only by utilizing the depth map from the multiple-view geometry or single-view depth estimation.